\pdfoutput=1
\documentclass[11pt]{article}

\usepackage[]{acl}

\usepackage{times}
\usepackage{latexsym}

\usepackage[T1]{fontenc}
\usepackage[utf8]{inputenc}

\usepackage{microtype}

\usepackage{inconsolata}

\usepackage{framed}
\usepackage{mathtools}
\usepackage{graphics}
\usepackage{graphicx}
\usepackage{amsmath}
\usepackage{amsfonts,amssymb}
\usepackage{amssymb}
\usepackage{bbm}
\usepackage[ruled, linesnumbered]{algorithm2e}
\usepackage{amsmath}
\usepackage{booktabs}
\usepackage{multirow}
\usepackage{color}
\usepackage{xcolor}
\usepackage{pifont}
\usepackage{pgfplots}
\usepackage{makecell}
\usepackage{caption}
\usepackage{subcaption}

\usepackage{arydshln}
\usepackage{float}

\title{Packing Analysis: Packing Is More Appropriate for Large Models or Datasets in Supervised Fine-tuning}

\author{Shuhe Wang$^{1}$, Guoyin Wang, Yizhong Wang$^{2}$\\ 
{\bf Jiwei Li$^{3}$, Eduard Hovy$^{1,4}$, Chen Guo$^{5}$}\\
$^1$The University of Melbourne, $^2$University of Washington\\ $^3$Zhejiang University, $^5$China Mobile Group Device Co., Ltd\\ $^4$Language Technologies Institute, Carnegie Mellon University}
\begin{document}
\maketitle
\begin{abstract}
\let\thefootnote\relax\footnotetext{Email: shuhewang@student.unimelb.edu.au}
Packing, initially utilized in the pre-training phase, is an optimization technique designed to maximize hardware resource efficiency by combining different training sequences to fit the model's maximum input length.
Although it has demonstrated effectiveness during pre-training, there remains a lack of comprehensive analysis for the supervised fine-tuning (SFT) stage on the following points:
(1) whether packing can effectively enhance training efficiency while maintaining performance, (2) the suitable size of the model and dataset for fine-tuning with the packing method, and (3) whether packing unrelated or related training samples might cause the model to either excessively disregard or over-rely on the context.

In this paper, we perform extensive comparisons between SFT methods using padding and packing, covering SFT datasets ranging from 69K to 1.2M and models from 8B to 70B.
This provides the first comprehensive analysis of the advantages and limitations of packing versus padding, as well as practical considerations for implementing packing in various training scenarios. 
Our analysis covers various benchmarks, including knowledge, reasoning, and coding, as well as GPT-based evaluations, time efficiency, and other fine-tuning parameters. We also open-source our code for fine-tuning and evaluation and provide checkpoints fine-tuned on datasets of different sizes, aiming to advance future research on packing methods.\footnote{Code is available at: \url{https://github.com/ShuheWang1998/Packing-Analysis?tab=readme-ov-file}}
\end{abstract}

\section{Introduction}
Supervised fine-tuning (SFT) refers to adapting a pre-trained model to perform specific tasks by training it on a labeled conversation dataset consisting of (instruction, answer) pairs \cite{ouyang2022training, muennighoff2022crosslingual, wang2022self, taori2023stanford, chiang2023vicuna, falcon40b, wang2023far, bai2023qwen, wang2023sim, cai2024internlm2, young2024yi}.
As models and datasets grow, the cost of fine-tuning rises. Identifying cost-effective methods, optimizing resource utilization, and alleviating the financial burden of large-scale training present new challenges in the SFT process \cite{hu2021lora, zhang2023adalora, dettmers2024qlora}.

Packing addresses these challenges by combining multiple training samples into a single sample. Originally used during the pre-training phase, packing extends each training sequence to the model's maximum input length, optimizing hardware resource usage, such as GPUs, and improving training efficiency \cite{brown2020language, rae2021scaling, chowdhery2022palm, openai2023gpt4, touvron2023llama, dubey2024llama}. 
Despite its proven effectiveness during the pre-training phase, for SFT, a thorough analysis is still lacking on: (1) whether packing can effectively enhance training efficiency while maintaining performance, (2) the suitable size of the model and dataset for fine-tuning with the packing method, and (3) whether packing unrelated or related training samples might cause the model to either excessively disregard or over-rely on the context.

To address these concerns, this paper provides a thorough analysis of packing during the supervised fine-tuning (SFT) stage.
Specifically, we perform extensive comparisons between supervised fine-tuning (SFT) methods using padding and packing, covering SFT datasets ranging from 69K to 1.2M and models from 8B to 70B. Our comparisons include various benchmarks, such as knowledge, reasoning, and coding, GPT-based evaluations, time efficiency, and other fine-tuning parameters, concluding that:

\begin{itemize}
	\item Models using packing generally perform better on average compared to those using padding across various benchmarks.
	\item As the model size grows, the performance gap between padding and packing-based models on the benchmark increases.
	\item Tailoring the packing of specific training samples may result in desired performance on specific benchmarks.
	\item Compared to padding, the packing method greatly reduces training time, making it possible to fine-tune large models on large datasets.
	\item Using longer training samples increases the time required for the packing method to process each sample, making it less suitable for training on particularly small datasets.
	\item In packing mode, the batch size is no longer directly proportional to the learning rate.
	\item Applying packing to datasets with only single-turn conversations may lead to a significant decrease in performance on few-shot benchmarks.
\end{itemize}

Building on these findings, we provide the first comprehensive analysis of the advantages and limitations of packing compared to padding, as well as practical considerations for implementing packing in various training scenarios. Additionally, we have open-sourced our code for fine-tuning and evaluation and released checkpoints fine-tuned on datasets of varying sizes, contributing to future research on packing methods.

\section{Related Work}
Supervised fine-tuning (SFT) in large language models (LLMs) involves additional training on a dataset of (instruction, answer) pairs. This approach helps align the LLMs’ training goal of predicting the next word with users’ expectations for the models to follow human instructions more accurately \cite{mishra2021cross, wei2021finetuned, rosenbaum2022linguist, ouyang2022training, wang2022super, dwivedi2022editeval, longpre2023flan, zhang2023instruction, qi2023fine, chung2024scaling, liu2024improved}.

The initial step in SFT is to create annotated data, but current SFT datasets are often constrained by their limited quantity, diversity, and creativity \cite{mukherjee2023orca, xu2023wizardlm, lu2023instag, song2024scaling, wang2023far, zhou2024lima}. To address this issue and rich resources for research, in one line, researchers distilled data from powerful large models (e.g., GPT-4 \cite{openai2023gpt4}) \cite{chiang2023vicuna, ding2023enhancing, zhao2024wildchat}. On the other line, some researchers are working on methods to enable pre-trained models to self-generate useful SFT data \cite{wang2022self, xu2023wizardlm, sun2024principle}. 

Once high-quality SFT datasets are created, the next step is to use them for fine-tuning pre-trained models. Many studies are dedicated to minimizing the costs of fine-tuning, including GPU usage and time, while maintaining performance, such as light-weight fine-tuning \cite{hu2021lora, zaken2021bitfit, zhang2023adalora, dettmers2024qlora}, speeding up attention algorithm for transformer-based LLMs \cite{dao2022flashattention}, and efficient distributed fine-tuning \cite{rajbhandari2020zero}\footnote{\url{https://github.com/microsoft/DeepSpeed}}. In this paper, we are analyzing one of the efficient fine-tuning techniques: packing, which packs multiple training samples into a single sample to maximize the utilization of hardware resources and enhance fine-tuning efficiency, providing a comprehensive understanding of its effectiveness and potential risks.

In this paper, we explore a new fine-tuning technique called packing, which gained prominence during the pre-training phase.
This method involves combining multiple training samples into one to optimize hardware resource usage and improve fine-tuning efficiency. Below, we will thoroughly analyze its effectiveness and potential risks.

\begin{figure*}[htb]
\centering
    \includegraphics[scale=0.4]{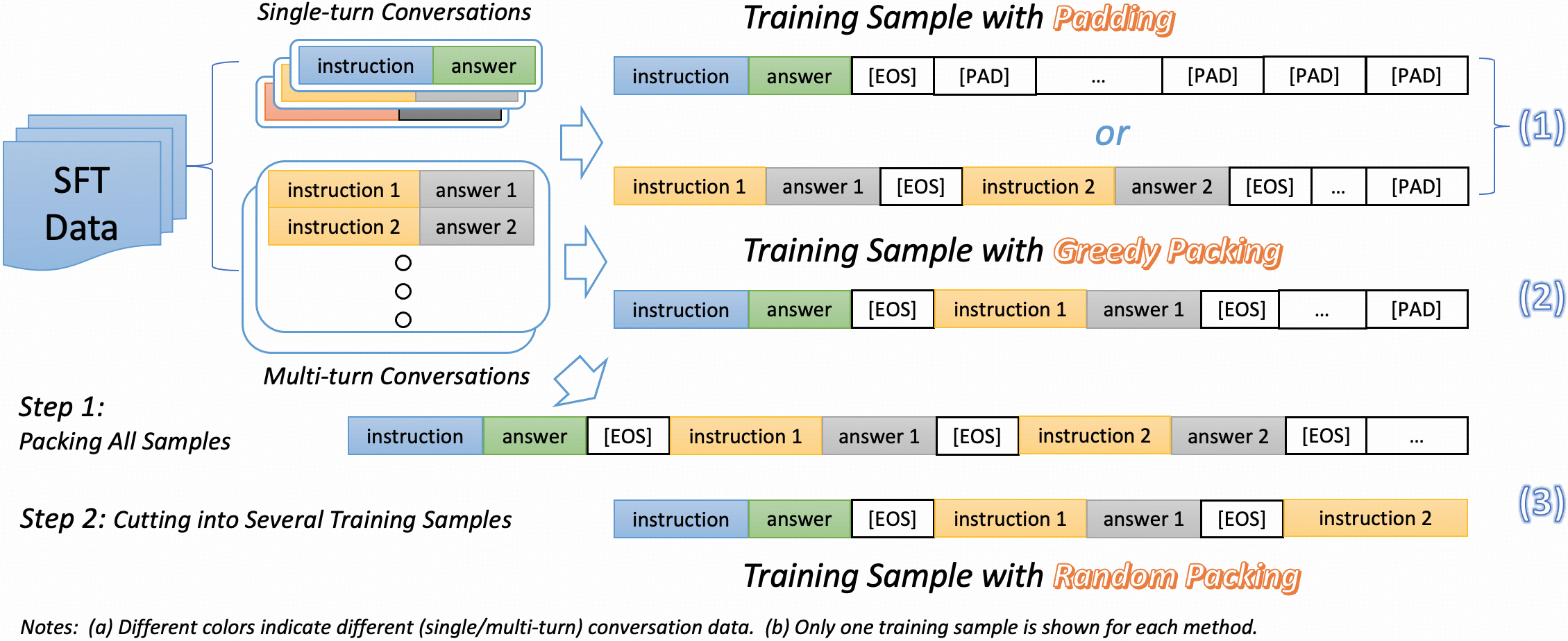}
  \caption{An example for the process of padding and packing methods: \textbf{(1) Padding}: Each training sample is appended with the special token “[PAD]” to meet the requirement of the model's input length; \textbf{(2) Greedy Packing}: Each training sample is packed together as much as possible according to its length; and \textbf{Random Packing}: firstly, all training samples are packed into one single sample, and then the single sample is cut into several short training samples according to the maximum input length of the model. It's important to note that random packing can sometimes split a single conversation across two different sequences, as illustrated by the conversation (instruction 2, answer 2) at the bottom of the figure.}
  \label{fig:methods}
\end{figure*}

\section{Methods: Padding and Packing}
Padding and packing are two distinct methods to organize training samples, shown in Figure \ref{fig:methods}. In this section, we first define a set of mathematical symbols, followed by a detailed explanation of one padding method and two packing methods: random packing and greedy packing.

\subsection{Symbols}
%\subsection{Background}

%Before detailing the analyzed fine-tuning methods, we use mathematical symbols to abstractly represent the data, models, and the supervised fine-tuning (SFT) process. 

We assume that $\{C\} = \{c^{1}, \dots ,c^{N}\}$ denotes the training conversations, where $N$ is the size of the training conversation, and $c^{i} = \{(x^{i}_{1}, y^{i}_{1}), \dots , (x^{i}_{m}, y^{i}_{m})\}, m \geq 1$ denotes an instruction $x$ and answer $y$ pair of length $m$. 
The SFT process tunes a pre-trained LLM $f$ to create a more human-like model $g$ that adheres to user instructions and human preferences. This typically consists of two main steps: (1) preprocessing the training conversation into a series of batches, and (2) feeding each batch into the pre-trained LLM $f$ and performing backpropagation to adjust the parameters of LLM $f$.
In the following paragraphs, we also use the terms "\textit{single-turn conversation}" and "\textit{multi-turn conversation}" to distinguish between training conversations that consist of a single instruction and answer pair ($m=1$) and those that consist of multiple contextual instruction and answer pairs ($m \geq 2$). 

%In this section, we focus on three distinct methods, shown in Figure \ref{fig:methods}, for batching training conversations: (1) Padding, which involves padding each training conversation with special token to the maximum input limit of the pre-trained model and then randomly dividing the full training conversations into $M_{pad}$ batches of size $l$; (2) Random Packing, which combines all training conversations in random order into one large multi-turn conversation, then cuts this combined conversation into $M^{random}_{pack}$ batches of size $l$ based on the maximum input limit of the pre-trained model; and (3) Greedy Packing, which greedily merging training conversations by their lengths to closely match the maximum input length of the pre-trained model, and then subsequently dividing the merged conversations into $M^{greedy}_{pack}$ batches of size $l$. 

%As for the way of backpropagation, these three methods are roughly the same, so we will cover them briefly below.

\subsection{Padding}
Padding refers to extending the length of training conversations to a consistent size to match the input requirements of the pre-trained LLM, following are its detailed process, strengths, and weaknesses:

\subsubsection{Process of Padding}
%\paragraph{Method.}
Specifically, for a training sample comprising $m$-turn ($m \geq 1$) conversations $\{(x_{1}, y_{1}),\dots ,(x_{m}, y_{m}) \}$, we start by concatenating each turn of the conversation into a single sequence, using the special token $[EOS]$ to separate each instruction $x$ and response $y$:

\begin{equation}
\nonumber
    \begin{aligned}
     	&x_{1} y_{1}  [EOS] x_{2} y_{2} [EOS]  \cdots  x_{m} y_{m} [EOS] \\
     	&= \\
     	&w_{1} w_{2} \cdots [EOS] w_{t} w_{t+1} \cdots w_{T} [EOS]
    \end{aligned}
\end{equation}
where $w_{i}, 1 \leq i \leq T$ denotes the word of the concatenated sequence, and $T$ denotes the number of words in this sequence.

During the training stage, LLMs often process data in batches, and each batch must consist of input sequences of equal length. This uniformity is essential because the fundamental mathematical operations required, such as matrix multiplications, necessitate tensors of consistent sizes. In this condition, for one batch with $l$ concatenated sequences:
\begin{equation}
\nonumber
    \begin{aligned}
     	&w^{1}_{1} w^{1}_{2} \cdots [EOS] w^{1}_{t_{1}} w^{1}_{t_{1}+1} \cdots w^{1}_{T_{1}} [EOS] \\
     	&w^{2}_{1} w^{2}_{2} \cdots [EOS] w^{2}_{t_{2}} w^{2}_{t_{2}+1} \cdots w^{2}_{T_{2}} [EOS] \\
     	&\cdots \\
     	&w^{l}_{1} w^{l}_{2} \cdots [EOS] w^{l}_{t_{l}} w^{l}_{t_{l}+1} \cdots w^{l}_{T_{l}} [EOS]
    \end{aligned}
\end{equation}
we must ensure that all sequences in this batch are limited to the minimum of the maximum input length of the pre-trained model (for example, 8192 for LLaMA-3-8B-base \cite{dubey2024llama}) and the length of the longest sequence within the batch:
\begin{equation}
\nonumber
    \begin{aligned}
     	T = \min(\text{Maximum Input Length}, \mathop{\arg\max}\limits_{1\leq i \leq l} T_{i})
    \end{aligned}
\end{equation}

Thus, for sequences that exceeding length $T$, we truncate them, and for those shorter than $T$, we fill them up with non-informative special token $[PAD]$ at their end:
\begin{equation}
\nonumber
    \begin{aligned}
     	&w^{1}_{1} w^{1}_{2} \cdots [EOS] w^{1}_{t_{1}} \cdots [EOS] \cdots w^{1}_{T} [EOS] \\
     	&w^{2}_{1} w^{2}_{2} \cdots [EOS] w^{2}_{T-2}\ \ [PAD][PAD][EOS] \\
     	&\cdots \\
     	&w^{l}_{1} w^{l}_{2} \cdots [EOS] w^{l}_{t_{l}} \cdots w^{l}_{T-1}\ \ [PAD][EOS]
    \end{aligned}
\end{equation}

\subsubsection{Strengths or Padding}
%\paragraph{Strengths.}
Padding mainly has two strengths, in one line, padding is straightforward to implement, allowing for its use across various training frameworks. In the other line, by truncating longer sequences and padding shorter ones, padding ensures that all training data conforms to the model's architecture, preventing errors associated with varying sequence lengths and speeding up underlying matrix computations.

\subsubsection{Weaknesses of Padding}
%\paragraph{Weaknesses.}
While necessary, excessive padding can lead to inefficiencies. If there are many padding tokens relative to actual data, it can lead to increased computation without corresponding benefits in learning. In order to alleviate this weakness, below, we introduce two packing-based methods: Random Packing and Greedy Packing.

\subsection{Random Packing}
Unlike the padding method, which extends shorter sequences with the special token $[PAD]$, random packing combines multiple training conversations into a single sequence randomly, to maximize the model's learning efficiency and effectiveness:

\subsubsection{Process of Random Packing}
%\paragraph{Method.} 
Firstly, we concatenate all training conversations $\{C\} = \{c^{1}, \dots ,c^{N}\}$ into one single sequence with the special token $[EOS]$ to separate each instruction $x$ and response $y$:

\begin{equation}
\nonumber
    \begin{aligned}
     	&c_{1} [EOS]  c_{2}  [EOS]  \cdots  [EOS] c_{N} [EOS] \\
     	&= \\
     	&x_{1} y_{1}  [EOS] x_{2} y_{2}  \cdots [EOS] x_{n}  y_{n} [EOS] \\
     	&= \\
     	&w_{1} w_{2} \cdots [EOS] w_{t} w_{t+1} \cdots w_{m} [EOS]
    \end{aligned}
\end{equation}
where $n$ represents the total number of training instructions $x$ and responses $y$, $w_{i}, i\leq i \leq m$ denotes each word in the concatenated sequence, and $m$ signifies the total number of words in that sequence.

Secondly, assuming that the maximum input length for the pre-trained model is $T$, we adjust to this limit by dividing the concatenated sequence into $M$ subsequences, each with a length of $T$:

\begin{equation}
\nonumber
    \begin{aligned}
     	&w^{1}_{1} w^{1}_{2} \cdots [EOS] w^{1}_{t} w^{1}_{t+1} \cdots w^{1}_{T} [EOS] \\
     	&w^{2}_{1} w^{2}_{2} \cdots [EOS] w^{2}_{t} w^{2}_{t+1} \cdots w^{2}_{T} [EOS] \\
     	&\cdots \\
     	&w^{M}_{1} w^{M}_{2} \cdots [EOS] w^{M}_{t} w^{M}_{t+1} \cdots w^{M}_{T^{'}} [EOS]
    \end{aligned}
\end{equation}
where $T^{'} \leq T$, the special $[PAD]$ token will be appended when $T^{'} < T$.

In the end, we randomly pack these sequences into batches with each size of $l$, forming like:
\begin{equation}
\nonumber
    \begin{aligned}
     	&w^{1}_{1} w^{1}_{2} \cdots [EOS] w^{1}_{t} w^{1}_{t+1} \cdots w^{1}_{T} [EOS] \\
     	&w^{2}_{1} w^{2}_{2} \cdots [EOS] w^{2}_{t} w^{2}_{t+1} \cdots w^{2}_{T} [EOS] \\
     	&\cdots \\
     	&w^{l}_{1} w^{l}_{2} \cdots [EOS] w^{l}_{t} w^{l}_{t+1} \cdots w^{l}_{T} [EOS]
    \end{aligned}
\end{equation}

\subsubsection{Strengths of Random Packing}
%\paragraph{Strengths.}
Compared to the padding method, random packing enhances computational efficiency by densely packing each training batch, minimizing unused space and optimizing the use of the model's capacity. Furthermore, this approach potentially boosts the model's ability to generalize by exposing it to a broader range of contextual combinations in each training sample, thereby providing a more diverse array of scenarios.

\subsubsection{Weaknesses of Random Packing}
%\paragraph{Weaknesses.}
%One potential weakness of randomly packing training samples is that 
There are two potential weaknesses of random packing, one is that it can lead to the concatenation of two distinct or similar samples, which may cause the model to either excessively ignore or rely on the context.
We have put more analyses about this issue in Section \ref{sec:results_and_analysis}, here, briefly, the packing method does not cause the model to focus on the packed context overly. There are primarily two reasons for this: (1) the likelihood that a random approach leads to two similar training samples being packed together is quite low, and even if this occurs, it represents a minor proportion of the entire training set, thus not significantly influencing the model's capabilities; (2) the special token $[EOS]$ effectively allows the model to differentiate between two adjacent training samples.

The other potential issue is that the method of combining all training conversations and then dividing them into sequences could result in a single conversation being split across two different sequences. For example, the instruction might end up at the tail end of one sequence, while the corresponding answer starts at the beginning of the next sequence, which is shown at the bottom of Figure \ref{fig:methods}. To alleviate this issue, in the following section, we turn to the other packing-based method: Greedy Packing.

\subsection{Greedy Packing}
Instead of random packing that might result in a single conversation being split across two different sequences, greedy packing starts by sorting and selecting training conversations based on their length:

\subsubsection{Process of Greedy Packing}
%\paragraph{Method.} 
Formally, for a $m$-turn ($m \geq 1$) training conversation $\{(x_{1}, y_{1}),\dots ,(x_{m}, y_{m}) \}$, we first use the special token $[EOS]$ to concatenate all instructions $x$ and answers $y$ into one single sequence $s$: 

\begin{equation}
\nonumber
    \begin{aligned}
     	&x_{1} y_{1}  [EOS]  \cdots  [EOS] x_{m} y_{m} [EOS] \\
     	&= \\
     	&w_{1} w_{2} \cdots [EOS] w_{t} w_{t+1} \cdots w_{T} [EOS]
    \end{aligned}
\end{equation}
where $w_{i}, 1 \leq i \leq T$ denotes the word of the concatenated sequence, and $T$ denotes the number of words in this sequence. 

Then, we sort all of the concatenated sequences $s$, iterating from the longest sequence, and, in a greedy way, we pack as many sequences as possible without exceeding the maximum input length allowed by the pre-trained model. The full process is present in Algorithm \ref{alg:greedy_packing}, which results in $M$ packed sequences $S$:
\begin{equation}
\nonumber
    \begin{aligned}
     	&w^{1}_{1} w^{1}_{2} \cdots [EOS] w^{1}_{t_{1}} w^{1}_{t_{1}+1} \cdots w^{1}_{T_{1}} [EOS] \\
     	&w^{2}_{1} w^{2}_{2} \cdots [EOS] w^{2}_{t_{2}} w^{2}_{t_{2}+1} \cdots w^{2}_{T_{2}} [EOS] \\
     	&\cdots \\
     	&w^{M}_{1} w^{M}_{2} \cdots [EOS] w^{M}_{t_{M}} w^{M}_{t_{M}+1} \cdots w^{M}_{T_{M}} [EOS]
    \end{aligned}
\end{equation}

Finally, similar to the padding method, we truncate packed sequences that exceed the maximum length allowed by the pre-trained model and pad shorter sequences with the special token $[PAD]$ to randomly batch the packed sequences.

\SetKwComment{Comment}{\% }{}
\begin{algorithm}
\caption{Greedy Packing Training Sequences.}\label{alg:greedy_packing}

\KwData{$s_{1},\dots,s_{N}$} \Comment{Concatenated training sequences}
\KwResult{$S_{1},\dots,S_{M}$} \Comment{Packed sequences}
\Comment{Sort training sequences based on their lengths}
SORT ($s_{1},\dots,s_{N}$) \\
\Comment{Initialize the max input length of the pre-trained model}
$MaxLength \gets$ The max input length of the pre-trained model \\
\Comment{Initialize the index of the packed sequence $S_{j}$}
$j \gets 1$ \\
\For{$i=N,\dots,1$}{ 
	\Comment{Skip the training sequence that has been visited}
	\If{$i$ has not been visited}{
		\eIf{length of $(S_{j}+s_{i})\leq MaxLength$}{
			\Comment{Pack the training sequence $s_{i}$ into $S_{j}$}
			$S_{j}\gets S_{j}+s_{i}$ \\
			\Comment{Mark the training sequence $i$ as the visited state}
			Visited $i$ 
		}{
			\If{length of $S_{j}\neq 0$}{
				\Comment{Skip the packed sequence that is longer than the max input length of the pre-trained model}
				$j\gets j + 1$
			}
			\Comment{Initialize the packed sequence $S_{j}$ with the training sequence $s_{i}$}
			$S_{j}\gets s_{i}$ \\
			Visited $i$
		}
	}

}
\end{algorithm}

\subsubsection{Strengths of Greedy Packing}
%\paragraph{Strengths.}
Greedy packing mainly serves as a modification of random packing, designed to reduce the risk of dividing relevant contexts across different batches. Simultaneously, it preserves the benefits of the packing method: enhancing the model's generalization capabilities by exposing it to a wider variety of contextual combinations within each training sample, thus encompassing a more diverse set of scenarios.

\subsubsection{Weaknesses of Greedy Packing}
%\paragraph{Weaknesses.}
In addition to the issue associated with the packing method: it may cause the model to either excessively ignore or rely on the context by packing two distinct or similar training samples into one sequence. Another potential concern is the break in the random sampling of training data. Since greedy packing entails sorting and organizing data prior to batching, it naturally diminishes the randomness in the distribution of sequences across batches. This can affect the diversity of data within each batch, as it is not entirely random but instead guided by the specific criteria (sequence length), for packing. 
However, despite these concerns in subjective analysis, our analysis and a series of experimental results in Section \ref{sec:results_and_analysis} have shown that using models trained with the greedy packing method does not result in any performance loss across various downstream benchmarks and GPT-based evaluations.

\begin{table*}[ht]
    \tiny
    \centering
    \resizebox{.88\textwidth}{!}{
    \begin{tabular}{lcc}\toprule
   		\textbf{Dataset} & \textbf{\# Instance} & \textbf{\# Generator} \\\midrule
        Aya \cite{singh2024aya} & 202K & Human \\
        ChatArena \cite{zheng2024judging} & 33K & Open LLMs \\
        LIMA \cite{zhou2024lima} & 1K & Human \\
        MetaMathQA \cite{yu2023metamath} & 395K & GPT \\
        No Robots \cite{no_robots} & 9.5K & Human \\
        ShareGPT \cite{chiang2023vicuna} & 114K & GPT \\
        UltraChat 200K \cite{ding2023enhancing} & 200K & GPT and Human \\
        WildChat (GPT-4) \cite{zhao2024wildchat} & 69k & GPT \\
        Evol-Instruct \cite{xu2023wizardlm} & 143K & GPT \\
        FLAN \cite{longpre2023flan} & 100K & Human-LLMs Mixtures \\
        Alpaca GPT-4 \cite{peng2023instruction} & 20K & GPT \\
        Code Alpaca \cite{codealpaca} & 20K & Model Self-generation \\
        OpenOrca \cite{OpenOrca} & 30K & GPT-4 \\\midrule
        Open-source 1M & 1.3M & Mixture \\\bottomrule
    \end{tabular}
    }
    \caption{Details of the collected Open-source 1M dataset.}
    \label{tab:open_source_1m}
\end{table*}

\section{Experimental Setups}
In this section, we sequentially describe our “Training Setups” in \ref{training_setups} and “Evaluation Setups” in \ref{evaluation_setups}. For results and analysis, we put them in Section \ref{sec:results_and_analysis}.

\subsection{Training Setups}
\label{training_setups}

\subsubsection{Training Datasets}
The development of packing methods was primarily aimed at maximizing hardware resource utilization and minimizing training duration. To demonstrate these training differences between packing and padding, we analyze four SFT datasets with different sizes:

\paragraph{WildChat (GPT-4).} 
WildChat \cite{zhao2024wildchat} is a corpus comprising roughly 652k real-world interactions between users and ChatGPT, noted for its variety of languages and diverse user prompts. This dataset was created by providing users free access to ChatGPT and GPT-4 in return for their consent to collect chat histories. WildChat (GPT-4) is the smallest dataset in our experiments consisting of approximately 69k real-world interactions, selected specifically to include interactions between users and the GPT-4 model.

\paragraph{TULU.}
TULU \cite{wang2023far} is a dataset consisting of around 326k conversations, sourced both from real-world interactions between users and open large LLMs as well as from manually annotated dialogues. As a synthetic dataset, TULU \cite{wang2023far} aims to combine the benefits of various open resources, enhancing the performance of models fine-tuned on it to deliver the highest general performance.

\paragraph{WildChat (Full).} 
WildChat (Full) includes the entire 652k training conversations from the WildChat \cite{zhao2024wildchat} corpus. Utilizing such a large dataset allows us to confirm that the performance differences between padding and packing methods are statistically significant and not merely random fluctuations. Additionally, it provides an opportunity to assess the scalability and consistency of the padding and packing methods as the dataset is processed over time.

\paragraph{Open-source 1M.}
The larger the dataset, the more reliable the conclusions that can be drawn from the experiments, particularly in terms of how each method handles memory and computational resources at varying scales. To facilitate this analysis, we create a large data mixture named "open-source 1M", which consists of approximately 1.2M conversations collected from several high-quality open resources such as ShareGPT \cite{chiang2023vicuna}, FLAN V2 \cite{longpre2023flan}, Alpaca \cite{taori2023stanford}, among others. A complete list of these resources is detailed in Table \ref{tab:open_source_1m}.

\subsubsection{Model Training Details}

\paragraph{Pre-trained Models.}
In this paper, our experiments predominantly utilize the LLaMA-3-8B and LLaMA-3-70B \cite{dubey2024llama} models, which are among the largest and most advanced pre-trained models currently accessible to the research community:

\begin{table}[H]
%    \tiny
    \centering
    \resizebox{.48\textwidth}{!}{
    \begin{tabular}{lcc}\toprule
    	\textbf{Base LM} & \textbf{\# Params} & \textbf{\# Pre-training Tokens} \\\midrule
    	\multirow{2}{*}{\bf LLaMA-3} & 8B & 15T \\
    	& 70B & 15T\\\bottomrule
        
    \end{tabular}
    }
\end{table}

\paragraph{Chat Template.}
Following \citet{dubey2024llama}, we format all datasets to follow a chat template to unify the varied styles and formats of the instruction datasets:

\begin{figure}[htb]
\centering
    \includegraphics[scale=0.365]{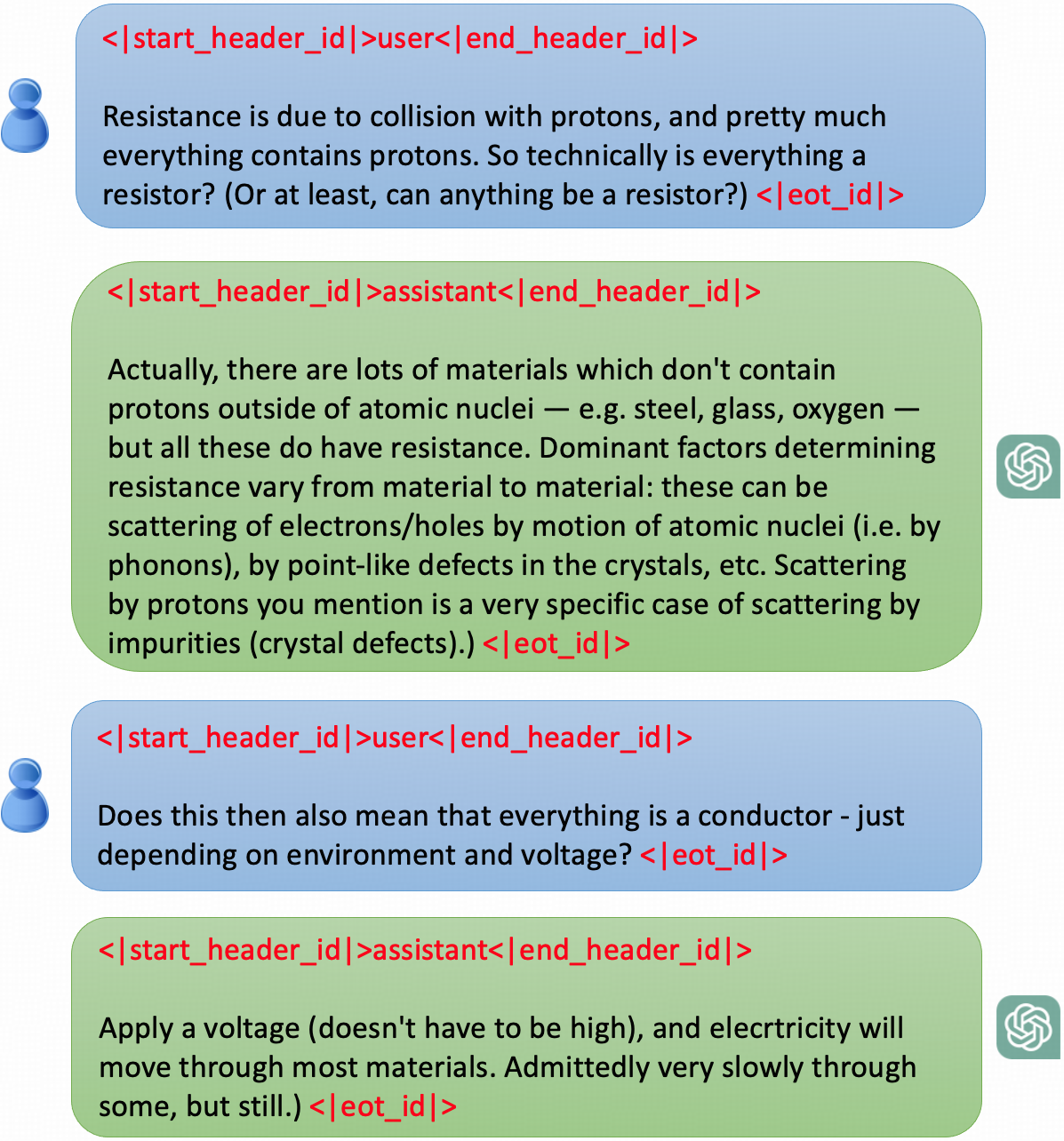}
  \label{fig:chat_template}
\end{figure}

As highlighted in red, we add special tokens “\text{<|start\_header\_id|>user<|end\_header\_id|>}” and “\text{<|start\_header\_id|>assistant<|end\_header\_id|>}” before instructions and answers separately, and an end flag “\text{<|eot\_id|>}” at the end of each instruction and answer, which at inference time, will guide the model to stop generating responses.

\paragraph{Training Details.}
During the training stage, we follow the default settings in \cite{wang2023far, dubey2024llama} masking loss belonging to the input (instruction), and only computing the loss after the special token “\text{<|start\_header\_id|>assistant<|end\_header\_id|>}”. All experiments were deployed the cluster with 4 nodes, each node containing 8 NVIDIA A800 80GB GPUs. For parameters, we keep the same between the padding and packing methods:
%, which are shown in Table \ref{tab:training_parameters}.

\begin{table}[H]
%    \tiny
    \centering
    \resizebox{.48\textwidth}{!}{
    \begin{tabular}{lcccc}\toprule
        & \multicolumn{2}{c}{{\bf LLaMA-3-8B}} & \multicolumn{2}{c}{{\bf LLaMA-3-70B}} \\
        & padding & packing & padding & packing \\\midrule
        Batch (per GPU) & 2 & 2 & 1 & 1 \\\midrule
        Gradient Acc & 2 & 2 & 2 & 2 \\\midrule
        Learning Rate & 1e-5 & 1e-5 & 1e-5 & 1e-5 \\\midrule
        Deepspeed & Stage-3 & Stage-3 & Stage-3 & Stage-3 \\\midrule
        Max Seq Length & 4096 & 4096 & 4096 & 4096 \\\midrule
        Warmup Ratio & 0.2 & 0.2 & 0.2 & 0.2 \\\midrule
        Epochs & 4 & 4 & 3 & 3 \\\midrule
        Offload Optimation & No & No & Yes & Yes \\\midrule
        Flash-Attention & \multirow{2}{*}{Yes} & \multirow{2}{*}{Yes} & \multirow{2}{*}{Yes} & \multirow{2}{*}{Yes} \\
        \cite{dao2022flashattention} &  &  &  & \\\bottomrule
    \end{tabular}
    }
    \caption{Training parameters for our experiments.}
    \label{tab:training_parameters}
\end{table}

\begin{table*}[ht]
%    \tiny
    \centering
    \resizebox{1.\textwidth}{!}{
    \begin{tabular}{lccccccc}\toprule
   		\textbf{Model} & \textbf{MMLU} & \textbf{GSM8K} & \textbf{MATH} & \textbf{BBH} & \textbf{IFEval} & \textbf{HumanEval} & \\
   		 & \textbf{Exact Match} & \textbf{Exact Match} & \textbf{Exact Match} & \textbf{Exact Match} & \textbf{(prompt-loose-accuracy)} & \textbf{pass@1} & \textbf{Avg} \\
   		 & \textbf{(5-shot)} & \textbf{(4-shot)} & \textbf{(4-shot)} & \textbf{(3-shot)} & \textbf{(0-shot)} & \textbf{(0-shot)} & \\\midrule
   		 \multicolumn{8}{c}{{\bf WildChat (GPT-4) Dataset, Size: 69K}} \\
   		 \multicolumn{8}{c}{{\it LLaMA-3-8B}} \\
   		 Padding & 63.99 & 58.76 & 14.72 & 60.71 & 56.01 & 43.29 & 49.58 \\
   		 Random Packing & 63.5\color{red}(-0.44) & \textbf{61.18\color{blue}(+2.42)} & 15.58\color{blue}(+0.86) & 61.04\color{blue}(+0.33) & 51.57\color{red}(-4.44) & \textbf{43.9 \color{blue}(+0.61)} & 49.46\color{red}(-0.12) \\
   		 Greedy Packing & \textbf{64.71\color{blue}(+0.72)} & 60.88\color{blue}(+2.12) & \textbf{15.6\color{blue}(+0.88)} & \textbf{62.59\color{blue}(+1.88)} & \textbf{57.12\color{blue}(+1.11)} & 42.68\color{red}(-0.61) & \textbf{50.6\color{blue}(+1.02)} \\\hdashline
   		 \multicolumn{8}{c}{{\it LLaMA-3-70B}} \\
   		 Padding & 73.47 & 79.3 & 28.8 & 78.33 & 51.76 & 57.32 & 61.50 \\
   		 Random Packing & \textbf{75.16\color{blue}(+1.69)} & \textbf{82.38\color{blue}(+3.08)} & 31.46\color{blue}(+2.66) & 79.94\color{blue}(+1.61) & 61.00\color{blue}(+9.24) & \textbf{65.85\color{blue}(+8.53)} & \textbf{65.97\color{blue}(+4.47)} \\
   		 Greedy Packing & 74.77\color{blue}(+1.3) & 81.61\color{blue}(+2.31) & \textbf{32.84\color{blue}(+4.04)} & \textbf{80.98\color{blue}(+2.65)} & \textbf{64.33\color{blue}(+12.57)} & 60.98\color{blue}(+3.66) & 65.92\color{blue}(+4.42) \\\midrule
   		 \multicolumn{8}{c}{{\bf TULU Dataset, Size: 326K}} \\
   		 \multicolumn{8}{c}{{\it LLaMA-3-8B}} \\
   		 Padding & 62.26 & 57.32 & 14.6 & 60.14 & 41.77 & 44.24 & 46.72 \\
   		 Random Packing & \textbf{63.94\color{blue}(+1.68)} & 58.83\color{blue}(+1.51) & 13.94\color{red}(-0.66) & 61.11\color{blue}(+0.97) & 42.51\color{blue}(+0.74) & \textbf{45.61(+1.37)} & 47.66\color{blue}(+0.94) \\
   		 Greedy Packing & 62.14\color{red}(-0.12) & \textbf{60.8\color{blue}(+3.48)} & \textbf{14.74\color{blue}(+0.14)} & \textbf{61.26(+1.12)} & \textbf{46.40\color{blue}(+4.63)} & 44.51\color{blue}(+0.27) & \textbf{48.31\color{blue}(+1.59)} \\\hdashline
   		 \multicolumn{8}{c}{{\it LLaMA-3-70B}} \\
   		 Padding & 73.2 & 81.18 & 29.02 & 78.06 & 47.32 & 62.95 & 61.96 \\
   		 Random Packing & \textbf{73.48\color{blue}(+0.28)} & \textbf{81.73\color{blue}(+0.55)} & 29.42\color{blue}(+0.4) & \textbf{78.35(+0.29)} & 47.29\color{red}(-0.03) & 60.37\color{red}(-2.58) & 61.77\color{red}(-0.19) \\
   		 Greedy Packing & 73.43\color{blue}(+0.23) & 81.2\color{blue}(+0.02) & 30\color{blue}(+0.18) & 77.54\color{red}(-0.52) & \textbf{53.05\color{blue}(+5.73)} & \textbf{68.9\color{blue}(+5.95)} & \textbf{64.02\color{blue}(+2.06)} \\\midrule
   		 \multicolumn{8}{c}{{\bf WildChat Dataset, Size: 652K}} \\
   		 \multicolumn{8}{c}{{\it LLaMA-3-8B}} \\
   		 Padding & 64.52 & 61.83 & 14.21 & 61.88 & 51.36 & 40.12 & 48.99 \\
   		 Random Packing & 64.46\color{red}(-0.06) & \textbf{62.77\color{blue}(+0.94)} & 14.44\color{blue}(+0.23) & 62\color{blue}(+0.12) & 50.28\color{red}(-1.08) & 40.24\color{blue}(+0.12) & 49.03\color{blue}(+0.04) \\
   		 Greedy Packing & \textbf{65.07\color{blue}(+0.55)} & 61.41\color{red}(-0.42) & \textbf{15.08(+0.87)} & \textbf{62.83\color{blue}(+0.95)} & \textbf{52.68\color{blue}(+1.32)} & \textbf{48.17(+8.05)} & \textbf{50.87\color{blue}(+1.88)} \\\hdashline
   		 \multicolumn{8}{c}{{\it LLaMA-3-70B}} \\
   		 Padding & 74.82 & 79.26 & 29.44 & 76.31 & 52.19 & 63.7 & 62.62 \\
   		 Random Packing & \textbf{75.67\color{blue}(+0.85)} & \textbf{80.1\color{blue}(+0.84)} & 30.37\color{blue}(+0.93) & 76.74\color{blue}(+0.43) & 52.43\color{blue}(+0.24) & \textbf{65.26\color{blue}(+1.56)} & 63.43\color{blue}(+0.81) \\
   		 Greedy Packing & 75.36\color{blue}(+0.46) & 79.45\color{blue}(+0.19) & \textbf{31.28\color{blue}(+1.84)} & \textbf{77.47\color{blue}(+1.16)} & \textbf{53.60\color{blue}(+1.41)} & 64.02\color{blue}(+0.32) & \textbf{63.53\color{blue}(+0.91)} \\\midrule
   		 \multicolumn{8}{c}{{\bf Open-source 1M Dataset, Size: 1.2M}} \\
   		 \multicolumn{8}{c}{{\it LLaMA-3-8B}} \\
   		 Padding & 63.7 & 77.08 & 27.96 & 63.45 & 48.39 & 45.22 & 54.3 \\
   		 Random Packing & \textbf{63.96\color{blue}(0.26)} & 77.26\color{blue}(+0.16) & \textbf{28.4\color{blue}(+0.44)} & \textbf{64.83\color{blue}(+1.38)} & 49.54\color{blue}(+1.15) & 45.73\color{blue}(+0.51) & 54.95\color{blue}(+0.65) \\
   		 Greedy Packing & 63.63\color{red}(-0.07) & \textbf{77.48\color{blue}(+0.4)} & 28.26\color{blue}(+0.3) & 63.01\color{red}(-0.44) & \textbf{51.57\color{blue}(+3.28)} & \textbf{46.34\color{blue}(+1.12)} & \textbf{55.05\color{blue}(+0.75)} \\\hdashline
   		 \multicolumn{8}{c}{{\it LLaMA-3-70B}} \\
   		 Padding & 74.97 & 85.23 & 41.82 & 78.65 & 54.33 & 61.74 & 66.12 \\
   		 Random Packing & \textbf{76.38\color{blue}(+1.41)} & 86.14\color{blue}(+0.91) & 42.73\color{blue}(+0.91) & 79.42\color{blue}(+0.77) & 55.9\color{blue}(+1.57) & \textbf{62.98\color{blue}(+1.24)} & 67.26\color{blue}(+1.14) \\
   		 Greedy Packing & 75.69\color{blue}(+0.72) & \textbf{86.88\color{blue}(+1.65)} & \textbf{42.92\color{blue}(+1.1)} & \textbf{79.94\color{blue}(+1.29)} & \textbf{56.82\color{blue}(+2.49)} & \textbf{62.98\color{blue}(+1.24)} & \textbf{67.54\color{blue}(+1.42)} \\\bottomrule
        \end{tabular}
    }
    \caption{Results of different size models and datasets on various benchmarks. We highlight the highest score in bold and use blue and red to indicate whether the score has increased or decreased compared to the padding method.}
    \label{tab:results_benchmarks}
\end{table*}

\begin{table*}[ht]
%    \tiny
    \centering
    \resizebox{1.\textwidth}{!}{
    \begin{tabular}{l|lcccc}\toprule
   		\multicolumn{2}{c}{{\bf Model}} & \textbf{WildChat (GPT-4), 69K} & \textbf{TULU, 326K} & \textbf{WildChat, 652K} & \textbf{Open-source 1M, 1.2M} \\\midrule
   		\multirow{3}{*}{LLaMA-3-8B} & \textit{padding} & 28.86 & 19.11 & 21.06 & 18.38 \\ 
   		 & \textit{random packing} & 27.89\color{red}(-0.97) & \textbf{20.84\color{blue}(+1.73)} & 20.73\color{red}(-0.33) & 20.42\color{blue}(+2.04) \\ 
   		 & \textit{greedy packing} & \textbf{29.81\color{blue}(+0.95)} & 20.73\color{blue}(+1.62) & \textbf{21.34\color{blue}(+0.28)} & \textbf{21.9\color{blue}(+3.52)} \\\midrule
   		 \multirow{3}{*}{LLaMA-3-70B} & \textit{padding} & 37.0 & 22.84 & 30.69 & 34.95 \\ 
   		 & \textit{random packing} & 39.92\color{blue}(+2.92) & 23.93\color{blue}(+1.09) & 30.76\color{blue}(+0.07) & 35.21\color{blue}(+0.26) \\ 
   		 & \textit{greedy packing} & \textbf{41.09\color{blue}(+4.09)} & \textbf{24.46\color{blue}(+1.62)} & \textbf{31.26\color{blue}(+0.57)} & \textbf{35.81\color{blue}(+0.86)} \\\bottomrule
        \end{tabular}
    }
    \caption{Results of different size models and datasets on the WildBench benchmark. We highlight the highest score in bold and use blue and red to indicate whether the score has increased or decreased compared to the padding method.
    Note that the WildChat (GPT-4) dataset is composed entirely of real user interactions with GPT-4, and its internal data is the foundation of WilBench research. Therefore, it is reasonable for a model trained on the WildChat dataset to achieve a high score.}
    \label{tab:wildbench_results}
\end{table*}

\begin{table*}[ht]
%    \tiny
    \centering
    \resizebox{.99\textwidth}{!}{
    \begin{tabular}{lccccc}\toprule
   		\textbf{Model} & \textbf{Epoch} & \textbf{Total Steps} & \textbf{Total Training Time (s)$\downarrow$} & \textbf{Steps per Second$\uparrow$} & \textbf{Samples per Second$\uparrow$} \\\midrule
   		 \multicolumn{6}{c}{{\bf WildChat (GPT-4) Dataset, Size: 69K}} \\
   		 \multicolumn{6}{c}{{\it LLaMA-3-8B}} \\
   		 Padding & 4 & 1964 & 1188.8449 & \textbf{0.165} & \textbf{21.13} \\
   		 Random Packing & 4 & 728 & 445.28773\color{blue}(-743.55717) & 0.163\color{red}(-0.002) & 20.934\color{red}(-0.196) \\
   		 Greedy Packing & 4 & 492 & \textbf{308.33346\color{blue}(-880.51144)} & 0.16\color{red}(-0.005) & 20.48\color{red}(-0.65) \\\hdashline
   		 \multicolumn{6}{c}{{\it LLaMA-3-70B}} \\
   		 Padding & 3 & 2943 & 9533.42936 & 0.031 & 1.976 \\
   		 Random Packing & 3 & 1092 & 3749.3016\color{blue}(-5784.12776) & 0.029\color{red}(-0.002) & 1.865\color{red}(-0.111) \\
   		 Greedy Packing & 3 & 741 & \textbf{2573.34781\color{blue}(-6960.08155)} & 0.029\color{red}(-0.002) & 1.84\color{red}(-0.136) \\\midrule
   		 \multicolumn{6}{c}{{\bf TULU Dataset, Size: 326K}} \\
   		 \multicolumn{6}{c}{{\it LLaMA-3-8B}} \\
   		 Padding & 4 & 9183 & 4906.59014 & \textbf{0.165} & \textbf{21.084} \\
   		 Random Packing & 4 & 1928 & \textbf{1175.43583\color{blue}(-3731.15431)} & 0.164\color{red}(-0.001) & 20.977\color{red}(-0.107) \\
   		 Greedy Packing & 4 & 1956 & 1328.12592\color{blue}(-3578.46422) & 0.147\color{red}(-0.018) & 18.841\color{red}(-2.243) \\\hdashline
   		 \multicolumn{6}{c}{{\it LLaMA-3-70B}} \\
   		 Padding & 3 & 13761 & 40735.40051 & \textbf{0.034} & \textbf{2.162} \\
   		 Random Packing & 3 & 2889 & \textbf{9758.68127\color{blue}(-30976.71924)} & 0.03\color{red}(-0.004) & 1.895\color{red}(-0.267) \\
   		 Greedy Packing & 3 & 2931 & 10313.89593\color{blue}(-30421.50458) & 0.028\color{red}(-0.006) & 1.82\color{red}(-0.342) \\\midrule
   		 \multicolumn{6}{c}{{\bf WildChat Dataset, Size: 652K}} \\
   		 \multicolumn{6}{c}{{\it LLaMA-3-8B}} \\
   		 Padding & 4 & 18340 & 11738.48881 & \textbf{0.156} & \textbf{20.183} \\
   		 Random Packing & 4 & 5348 & 3422.97918\color{blue}(-8315.50963) & \textbf{0.156} & 20.006\color{red}(-0.177) \\
   		 Greedy Packing & 4 & 4780 & \textbf{3124.28736\color{blue}(-8614.20145)} & 0.153\color{red}(-0.003) & 19.58\color{red}(-0.603) \\\hdashline
   		 \multicolumn{6}{c}{{\it LLaMA-3-70B}} \\
   		 Padding & 3 & 27510 & 97893.95669 & \textbf{0.034} & \textbf{2.261} \\
   		 Random Packing & 3 & 8025 & 28904.78592\color{blue}(-68989.17077) & 0.030\color{red}(-0.004) & 2.083\color{red}(-0.178) \\
   		 Greedy Packing & 3 & 7170 & \textbf{25124.6234\color{blue}(-72769.33329)} & 0.029\color{red}(-0.005) & 1.826\color{red}(-0.435) \\\midrule
   		 \multicolumn{6}{c}{{\bf Open-source 1M Dataset, Size: 1.2M}} \\
   		 \multicolumn{6}{c}{{\it LLaMA-3-8B}} \\
   		 Padding & 4 & 33064 & 19918.48664 & \textbf{0.168} & \textbf{21.413} \\
   		 Random Packing & 4 & 5400 & 3253.07972\color{blue}(-16665.40692) & 0.166\color{red}(-0.002) & 21.255\color{red}(-0.158) \\
   		 Greedy Packing & 4 & 5104 & \textbf{3175.09395\color{blue}(-16743.39269)} & 0.161\color{red}(-0.007) & 20.571\color{red}(-0.842) \\\hdashline
   		 \multicolumn{6}{c}{{\it LLaMA-3-70B}} \\
   		 Padding & 3 & 49596 & 184709.04470 & \textbf{0.031} & \textbf{2.306} \\
   		 Random Packing & 3 & 8103 & 29893.65963\color{blue}(-154815.38507) & 0.03\color{red}(-0.001) & 2.193\color{red}(-0.113) \\
   		 Greedy Packing & 3 & 7653 & \textbf{27426.66515\color{blue}(-157282.37955)} & 0.028\color{red}(-0.003) & 1.786\color{red}(-0.52) \\\bottomrule
        \end{tabular}
    }
    \caption{The training time of models across various datasets, with blue indicating an improvement over the padding method, while red represents a decrease in performance compared to the padding method.}
    \label{tab:time_performance}
\end{table*}

\subsection{Evaluation Setups}
\label{evaluation_setups}

Following the argument in \cite{wang2023far} that general-purpose models should be able to perform some core tasks before they can generalize to satisfy various practical needs, we first assess the core capabilities of our fine-tuned models using a set of specific benchmarks. Subsequently, we employ evaluations based on GPT-4 to gauge their proficiency in following instructions and aligning with human preferences.

\subsubsection{Specific Benchmarks}
We evaluate our models on the following benchmarks:
\paragraph{MMLU.} 
Massive Multitask Language Understanding (MMLU) \cite{hendrycks2020measuring} consists of 14079 questions covering 57 tasks including elementary mathematics, US history, computer science, law, and more. The wide range of subjects and complex questions make MMLU suitable for testing the model's language comprehension and decision-making capabilities.

\paragraph{MATH and GSM8K.}
MATH \cite{hendrycks2021measuring} and GSM8K \cite{cobbe2021training} are two distinct mathematical datasets utilized for evaluating different aspects of model capabilities. The MATH \cite{hendrycks2021measuring} dataset comprises 12,500 complex competition-level mathematics problems, primarily designed to assess the ability of models to tackle challenging and advanced mathematical questions typically encountered at the college level. Conversely, the GSM8K \cite{cobbe2021training} dataset contains 8,500 high-quality elementary school math problems, aimed at testing the basic mathematical reasoning abilities of models.

\paragraph{BBH.}
BBH, short for BIG-Bench Hard \cite{suzgun2022challenging}, is a subset of the BIG-Bench \cite{srivastava2022beyond} dataset comprising 23 challenging tasks. These tasks were selected because they consistently proved too difficult for current large language models to handle effectively. Requiring complex, multi-step reasoning, the BBH dataset is primarily utilized to assess the general reasoning capabilities of models, testing their ability to navigate and solve intricate problems.

\paragraph{HumanEval.}
HumanEval \cite{chen2021evaluating} consists of 164 programming problems, including language comprehension, algorithms, and simple mathematics, with some comparable to simple software interview questions. 
The primary purpose of this dataset is to assess the ability of models to generate correct programs based on provided docstrings.

\paragraph{IFEval.}
IFEval \cite{zhou2023instruction} consists of 500 prompts, each containing specific instructions like "write an article with more than 800 words" or "enclose your response in double quotation marks." This dataset is used to test the ability of large language models to accurately follow given instructions.

\subsubsection{Evaluations Based on GPT-4}
While human-based evaluation provides important insights into user preferences, it suffers from significant drawbacks like high labor costs and lack of real-time feedback. 
To address these limitations, we employ WildBench \cite{lin2024wildbench}, an automated evaluation framework based on GPT-4. 
WildBench consists of 1,024 tasks manually selected from over one million human-chatbot conversation logs.  
It employs advanced LLMs (e.g., GPT-4-turbo) alongside specifically tailored checklists to systematically evaluate models' outputs and provide structured explanations supporting scoring and comparisons. 

For settings, we use WildBench-v2\footnote{To safeguard against the leakage of test data, WildBench periodically releases new versions of its test set. WildBench-v2 is the version in: \url{https://huggingface.co/datasets/WildEval/WildBench-V2}} as the test set and gpt-4o-2024-05-13\footnote{Most advanced GPT-4 model, which can be found in: \url{https://platform.openai.com/docs/models/gpt-4o}} as the annotator. We use the OpenCompass toolkit \cite{2023opencompass}, which is a one-stop platform for large model evaluations, and official prompts to make sure our results can be comparable to those on the open leaderboard\footnote{\url{https://huggingface.co/spaces/allenai/WildBench}}.

\section{Results and Analysis}
\label{sec:results_and_analysis}

In this section, we provide our experimental results as well as analysis based on them.

\subsection{Analysis on Various Benchmarks}
Table \ref{tab:results_benchmarks} and Table \ref{tab:wildbench_results} show results of different size models and datasets on various benchmarks, from that we can observe that:

\textbf{(1) Models using packing generally perform better on average compared to those using padding across various benchmarks.} Compared to the padding method, the packing method exposes models to a wider variety of contextual combinations within each training sample, offering a more diverse set of scenarios and enhancing the models' ability to generalize. For example, 61.50 (Padding) v.s. 65.97 (Random Packing) on the model LLaMA-3-70B for the WildChat (GPT-4) dataset.

\textbf{(2) As the model size grows, the performance gap between padding and packing-based models on the benchmark increases.} This is due to enhanced contextual efficiency. As the model size increases, its ability to effectively utilize extended contexts improves, thereby magnifying the advantages of the diverse contextual combinations brought by packing. For example, on the WildChat (GPT-4) dataset, the average score is 49.58 (Padding) v.s. 50.6 (Greedy Packing) on the model LLaMA-3-8B, while on the model LLaMA-3-70B, it is amplified to 61.50 (Padding) v.s. 65.92 (Greedy Packing).

\textbf{(3) Models that use greedy packing generally perform better than those employing random packing across most benchmarks, with particularly strong results on the IFEVal and WildBench benchmarks, which assess instruction-following capabilities.} 
This advantage of the greedy packing method lies in its ability to maintain the coherence of multi-turn conversations, unlike random packing which may split such conversations across different training samples. By preserving the integrity of multi-turn conversations, greedy packing helps models better learn when to use prior context effectively. This improves the model's ability to selectively access relevant information in instructions, resulting in enhanced performance on benchmarks such as IFEVal and WildBench that evaluate instruction-following capabilities, for example, the IFEval score improved from 50.28 to 52.68 for the WildChat dataset based on the model LLaMA-3-8B, and 49.54 to 51.57 for the Open-source 1M dataset based on the model LLaMA-3-8B. This interesting observation also suggests a new direction: \textbf{tailoring the packing of specific training samples to achieve desired performance}, where we will put more effort in the future.

\subsection{Analysis on Training Time}
Table \ref{tab:time_performance} shows the training time of different size models on various training datasets, where we can find that:

\textbf{(1) Compared to padding, the packing method greatly reduces training time, making it possible to fine-tune large models on large datasets.}
The packing method significantly decreases training time by efficiently utilizing the available computational resources, for example, 9533s (Padding) v.s. 3749s (Random Packing) on the model LLaMA-3-70B for the WildChat (GPT-4) dataset, and 40735s (Padding) v.s. 9758s (Random Packing) on the model LLaMA-3-70B for the TULU dataset. This reduction is particularly beneficial for scaling up model training and enables the effective handling of larger models and more extensive datasets, for example, from 97893s (Padding) significantly dropping to 25124s (Greedy Packing) on the model LLaMA-3-70B for the WildChat dataset, and from 184709s (Padding) significantly dropping to 27426s (Greedy Packing) on the model LLaMA-3-70B for the Open-source 1M dataset, thus enhancing the overall training throughput and allowing for more complex and comprehensive model fine-tuning.

\textbf{(2) Using longer training samples increases the time required for the packing method to process each sample, making it less suitable for training on particularly small datasets.}
Compared to padding, the packing method results in a lower number of samples processed per second, for example, 21.13 (Padding) v.s. 20.48 (Greedy Packing) for the WildChat (GPT-4) dataset based on the model LLaMA-3-8B, and 2.162 (Padding) v.s. 1.895 (Random Packing) for the TULU dataset based on the model LLaMA-3-70B.
Therefore, if your goal is to fine-tune a small model (e.g., 6B, 8B, or 9B) on a small dataset (e.g., 20K or 30K), using the padding method might be more time-efficient.

\begin{figure}[htb]
\centering
    \includegraphics[scale=0.44]{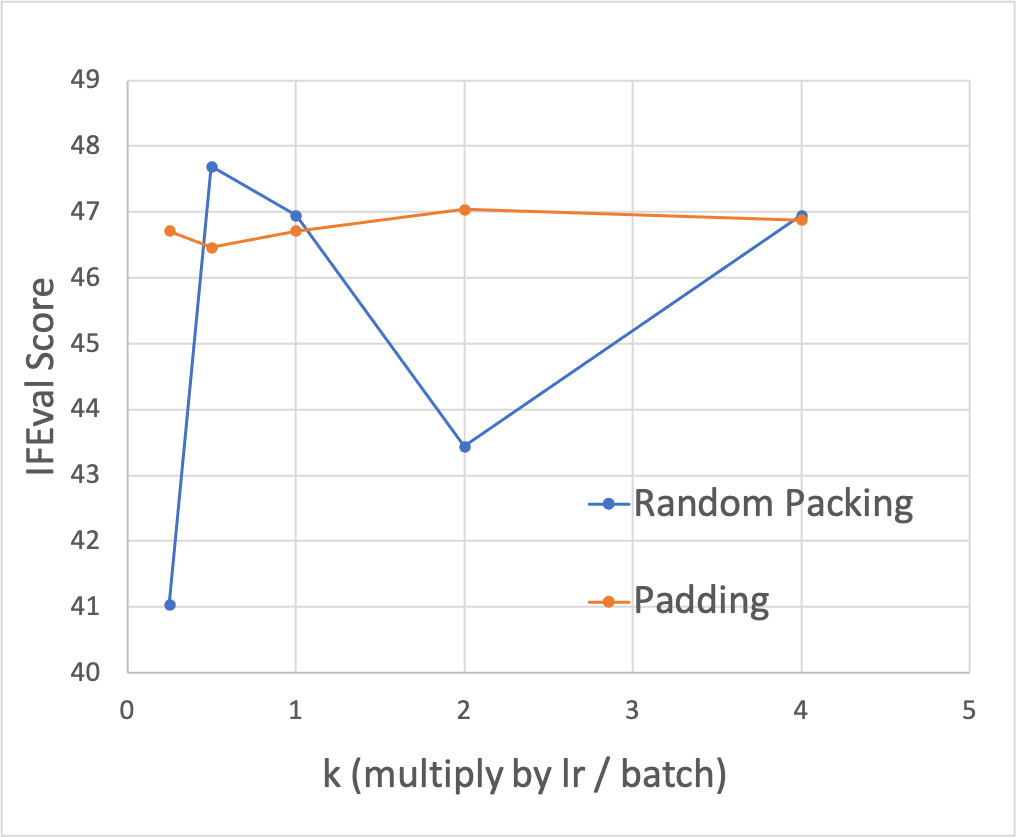}
  \caption{The results of fine-tuning the LLaMA-3-8B model on the TULU dataset using different linear combinations of batch size and learning rate.}
  \label{fig:linear_batch_lr}
\end{figure}

\subsection{Other Analysis}
In addition to the analysis provided, we conducted additional experiments and concluded that:

\textbf{(1) In packing mode, the batch size is no longer directly proportional to the learning rate.}
Previous research indicates that when increasing the batch size by a factor of $k$, the learning rate should also be multiplied by $k$ to maintain a constant variance in the gradient expectation \cite{goyal2017accurate}. This raises the question of whether the linear relationship between batch size and learning rate still holds when using the packing method. 
To determine this, we compare the padding method and the random packing method by separately fine-tuning the LLaMA-3-8B model on the TULU dataset using different linear combinations of batch size and learning rate. 
Results are shown in Figure \ref{fig:linear_batch_lr}, where the IFEval \cite{zhou2023instruction} score is the primary evaluation metric. 
The results reveal that while the batch size and learning rate adhere to a linear relationship in the padding method, this is not the case with the packing method. This discrepancy is due to the nature of packing: it does not ensure that each training sample consistently contains the same number of training conversations. Consequently, when the batch size is increased by a factor of $k$, the actual number of training conversations is not necessarily scaled up by the same factor, disrupting the linear relationship between batch size and learning rate.

\begin{figure}[htb]
\centering
    \includegraphics[scale=0.37]{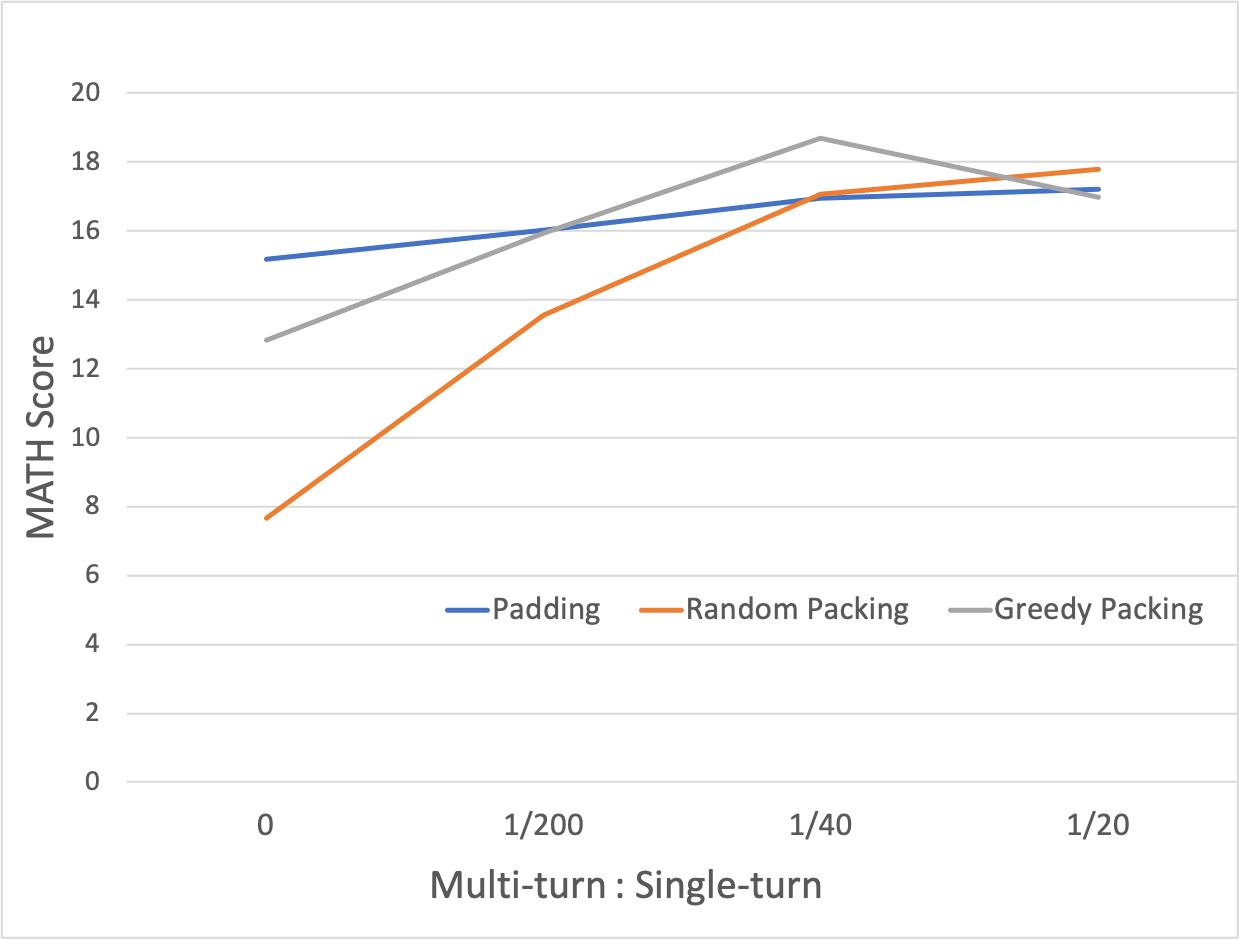}
  \caption{The results of fine-tuning the LLaMA-3-8B model by varying the ratio of multi-turn conversations and single-turn conversations.}
  \label{fig:single_turn_conversation}
\end{figure}

\textbf{(2) Applying packing to datasets with only single-turn conversations may lead to a significant decrease in performance on few-shot benchmarks.}
In packing methods, training samples that lack contextual connections may be combined to create what could be considered "fake" multi-turn conversations. When the training dataset includes multi-turn conversations, this approach allows the model to learn when to consider the context and when not to. However, if the training dataset only consists of single-turn conversations, there's a risk that the model might become less effective at utilizing context, potentially leading to a decline in performance on few-shot benchmarks.
To investigate this, we fine-tuned the LLaMA-3-8B model separately using the packing and padding methods on the filtered 200K OpenHermes 2.5 dataset\footnote{\url{https://huggingface.co/datasets/teknium/OpenHermes-2.5}}, which only consists of single-turn conversations.
The results, shown in Figure \ref{fig:single_turn_conversation}, reveal a significant drop in performance on the MATH \cite{hendrycks2021measuring} benchmark. Then, when we added multi-turn conversations into the fine-tuning dataset, scaling from $1/40$ to $1/20$ of multi-turn conversations was sufficient to restore performance to normal levels.

Interestingly, our recent experiments with our internal 200K high-quality single-turn dataset did not show any decline in performance on few-shot benchmarks. We believe this may be due to differences in data quality and plan to conduct further analysis. In the meantime, if you observe a significant drop in performance on few-shot benchmarks when using packing with datasets containing only single-turn conversations, consider incorporating a small amount of multi-turn conversations as a potential remedy.

\section{Conclusion}
In this paper, we conduct a thorough comparison of SFT methods using padding and packing, analyzing datasets from 69K to 1.2M and models ranging from 8B to 70B.
This provides the first detailed examination of the advantages and limitations of packing versus padding, as well as practical considerations for implementing packing in various training scenarios. 
Our evaluation spans a range of benchmarks, including knowledge, reasoning, and coding, and includes GPT-based assessments, time efficiency, and other fine-tuning factors. We also open-source our code for fine-tuning and evaluation, along with checkpoints fine-tuned on datasets of varying sizes, to support future research into packing techniques.

\bibliography{anthology,custom}

\appendix

\end{document}